\documentclass[journal]{vgtc}                     

\onlineid{0}
\usepackage{color, xcolor}           
\usepackage{array, colortbl}         
\usepackage{amsmath}
\usepackage{changes}
\definecolor{color1}{RGB}{248,149,064} 
\definecolor{color2}{RGB}{253,196,039} 
\definecolor{color3}{RGB}{252,254,164} 

\vgtccategory{Research}

\vgtcpapertype{please specify}

\title{DEGS: Deformable Event-based 3D Gaussian Splatting from RGB and Event Stream}

\author{%
  Junhao He$^{1\diamondsuit}$, Jiaxu Wang$^{1\diamondsuit}$, Jia Li, Mingyuan Sun, Qiang Zhang, Jiahang Cao, Ziyi Zhang, Yi Gu, Jingkai Sun, \\ and Renjing Xu$^{1\dagger}$
}
\authorfooter{
  \item 
  $^\dagger$Corresponding authors; $\diamondsuit$Co-first authors
  
  \item
  	Junhao He is with the Hong Kong University of Science and Technology (Guangzhou). 
  	E-mail: jhe382@connect.hkust-gz.edu.cn
  \item
        Jiaxu Wang is with the Hong Kong University of Science and Technology and the Hong Kong University of Science and Technology (Guangzhou). 
  	E-mail: jwang457@connect.ust.hk, jwang457@connect.hkust-gz.edu.cn
  \item
  	Jia Li is with the University of Hong Kong. 
  	E-mail: Jia.G.Lee@connect.hku.hk
  \item
  	Mingyuan Sun is with Northeastern University. 
  	E-mail: mingyuansun20@gmail.com
  \item
  	Qiang Zhang is with the Hong Kong University of Science and Technology (Guangzhou). 
  	E-mail: qzhang749@connect.hkust-gz.edu.cn
  \item
  	Jiahang Cao is with the Hong Kong University of Science and Technology (Guangzhou). 
  	E-mail: jcao248@connect.hkust-gz.edu.cn
  \item
  	Ziyi Zhang is with the Hong Kong University of Science and Technology (Guangzhou). 
  	E-mail: ziyizhang@hkust-gz.edu.cn
  \item
  	Yi Gu is with the Hong Kong University of Science and Technology (Guangzhou). 
  	E-mail: ygu425@connect.hkust-gz.edu.cn
  \item
  	Jingkai Sun is with the Hong Kong University of Science and Technology (Guangzhou). 
  	E-mail: jsun444@connect.hkust-gz.edu.com
  \item
        Renjing Xu is with the Hong Kong University of Science and Technology (Guangzhou). 
  	E-mail: renjingxu@hkust-gz.edu.cn
}

\abstract{%
Reconstructing Dynamic 3D Gaussian Splatting (3DGS) from low-framerate RGB videos is challenging. This is because large inter-frame motions will increase the uncertainty of the solution space. For example, one pixel in the first frame might have more choices to reach the corresponding pixel in the second frame. Event cameras can asynchronously capture rapid visual changes and are robust to motion blur, but they do not provide color information. Intuitively, the event stream can provide deterministic constraints for the inter-frame large motion by the event trajectories. Hence, combining low-temporal resolution images with high-framerate event streams can address this challenge. 
However, it is challenging to jointly optimize Dynamic 3DGS using both RGB and event modalities due to the significant discrepancy between these two data modalities. This paper introduces a novel framework that jointly optimizes dynamic 3DGS from the two modalities. The key idea is to adopt event motion priors to guide the optimization of the deformation fields. First, we extract the motion priors encoded in event streams by using the proposed LoCM unsupervised fine-tuning framework to adapt an event flow estimator to a certain unseen scene. Then, we present the geometry-aware data association method to build the event-Gaussian motion correspondence, which is the primary foundation of the pipeline, accompanied by two useful strategies, namely motion decomposition and inter-frame pseudo-label. Extensive experiments show that our method outperforms existing image and event-based approaches across synthetic and real scenes and prove that our method can effectively optimize dynamic 3DGS with the help of event data. 
}


\graphicspath{{figs/}{figures/}{pictures/}{images/}{./}} 

\usepackage{tabu}                      
\usepackage{booktabs}                  
\usepackage{lipsum}                    
\usepackage{mwe}                       

\usepackage{mathptmx}                  

\begin{document}



\firstsection{Introduction}

\maketitle

\label{sec.s intro}
High-quality dynamic scene reconstruction from a monocular video is significant for various applications, such as AR/VR, animation modeling, computer graphics, and 3D content creation. Previous approaches \cite{cao2023hexplane,gao2021dynamic,guo2022neural,liu2023robust,pumarola2021d,wu20234d,yang2023real} extend the conventional static reconstruction methods, such as Neural Radiance Field (NeRF) and 3D Gaussian Splatting (3DGS), into temporal dimensions. These methods require the spatial and temporal dense input to model the dynamical process. Therefore, they cannot faithfully reconstruct dynamic scenes when the number of video frames is insufficient and sparse, which might be caused by the fast-deforming objects or low frame rate RGB cameras. 

The event camera, also known as the dynamic vision sensor (DVS), works very differently from traditional cameras. Instead of capturing full frames at regular intervals, the event camera only detects relative changes in brightness for each pixel independently and asynchronously. This allows it to capture rich information about objects' movements and deformations, leading to several advantages including high temporal resolution, high dynamic range, and reduced data storage. Therefore, the event camera can handle the challenges mentioned above. Pure event streams do not contain absolute radiance information. Fortunately, conventional event cameras generally have an APS sensor, which captures RGB images at a lower frame rate. In this case, the event stream and RGB images complement each other; the event stream provides rich information on relative changes between frames, while the sparse RGB images provide absolute color information. This work proposes to combine the two modalities for dynamic scene reconstructions. 

Previous studies \cite{hwang2023evnerf,wang2024physical} attempt to reconstruct 3D static scenes from pure event streams by incorporating the linearized event generation model into the NeRF pipeline. However, they cannot model dynamic scenes or optimize NeRF from the two modalities. Hence, they can only synthesize gray-scale static novel views. Very recently, DE-NeRF \cite{ma2023deformable} first adopted event and RGB data to optimize the dynamic NeRF. This method asynchronously estimates per-event color by using RGB and the event generation model and then uses these estimated colors along with sparse RGB images to jointly optimize the dynamic NeRF. Nevertheless, it fails to explore the rich inter-frame motion information in the event stream, resulting in a suboptimal reconstruction. 
On the other hand, even though NeRF delivers good results in neural 3D reconstruction, the original NeRF suffers from large training and rendering costs. Recent 3DGS \cite{kerbl3Dgaussians} significantly boosts the rendering speed to a real-time level by replacing the cumbersome volume rendering in NeRF with efficient differentiable splatting. Moreover, 3DGS can also produce higher-fidelity rendering results. Very recently, Deform-GS\cite{yang2023deformable} and 4DGS\cite{wu20234d} equipped 3DGS with a deformation field to model dynamic scenarios. However, these approaches still rely on the high framerate of image sequences. When the frame rate of the video is low, the motion between images is too large, which can lead to the inability to optimize the Gaussian motion trajectory modeled by the deformation field. Moreover, thanks to the high temporal rate of the event camera, our method is naturally robust to motion blur in RGB images, and can complement additional sharp edge information for those blurry scenario reconstruction.

In this work, we propose a novel framework that can efficiently optimize the Dynamic 3D Gaussian Splatting from the two modalities, namely event stream and sparse RGB images. The primary obstacle is the large motion between sparse images, which prevents the deformation field from being easily optimized. The key idea of this work is that the event trajectories on the 2D plane can be used to optimize the deformation field in the dynamic 3DGS because the edge motion of objects triggers events. First, we present LoCM framework, which combines low‑rank adaptation (LoRA) and Contrast Maximization to finetune an event flow predictor, ensuring alignment with out-of-distribution scenes while reserving the motion prior knowledge learned from large-scale pretraining. Second, we propose a geometry-aware method to build event-Gaussian data associations, which constitutes the computational backbone of the proposed optimization scheme. Next, we adopt the motion decomposition and inter-frame pseudo-label strategies to guide the optimization between consecutive images. 

Our main contributions can be summarized as follows:

\begin{itemize}
    \item We propose the LoCM unsupervised finetuning framework by leveraging low-rank adaptation and contrast maximization to adapt a pretrained event flow estimator to an unseen scenario while maintaining its original priors, ensuring the extraction of high-quality motion information from event data. 
    
    \item To take full advantage of the motion cues in the event stream, we propose the geometry-aware data association method that can build motion correspondence between 2D events and 3D Gaussians to utilize event trajectories to optimize the deformation field. 

    \item To mitigate the dynamic ambiguity, we propose to use a motion decomposition scheme and inter-frame pseudo labels to optimize the deformation between consecutive frames. 
  
    \item We establish a novel synthetic event-based dataset and we build a pipeline to convert commonly used image-based real-world datasets into the event-based version by using a real event camera DVXplorer\cite{Inivation} to facilitate the community. Experiments prove the effectiveness of optimizing dynamic scenes by the two modalities. 
\end{itemize}

\section{Related Work}
\label{sec: related work}
\subsection{Dynamic Novel View Synthesis}
Synthesizing novel views of a dynamic scenario from captured 2D sequences remains a challenge. Since NeRF\cite{mildenhall2021nerf} has achieved great success in novel view synthesis, many efforts have been made to generalize NeRF to capture dynamic scenes. Some attempts\cite{pumarola2021d, du2021neural, li2021neural, liu2023robust, park2021nerfies} combine NeRF with additional time dimension or time-conditioned latent codes to reconstruct time-varying scenarios. Some other works \cite{Fang_2022, guo2023forward, yi2023generalizable, fridovich2023k, gan2023v4d, li2022streaming, shao2023tensor4d, wang2023mixed} explicitly incorporate voxel grids to model temporal information. Additionally, some researchers \cite{cao2023hexplane, song2023nerfplayer, abouchakra2023particlenerf} attempt to construct explicit structures to learn a 6D hyperplane function without directly modeling. 

Recently, a novel point-based representation, i.e. 3DGS, has been presented which formulates points as 3D Gaussians with learnable parameters. Although the vanilla 3DGS regards the scene as static, a few works have attempted to extend 3DGS to dynamic scenes due to its real-time rendering and high reconstruction quality. D-3DGS \cite{luiten2023dynamic} is the first attempt to adopt 3DGS to dynamic scenes. Inspired by dynamic NeRFs, some works \cite{wu20234d, yang2023deformable, yang2023real,yan2024street} introduce the deformed-based 3DGS that preserves a set of canonical Gaussians and learns the deformation field at each timestep. These works introduce the topological invariance into the training pipeline, thereby enhancing their suitability for the reconstruction of dynamic scenarios from monocular inputs. Besides, some other works \cite{xie2023physgaussian, guo2024motion, feng2024gaussian, zhong2024reconstruction} choose to explicitly formulate the continuous motion for deformation using the assumption that the dynamics of the scene are the consequences of the movement. In this work, our method proposes to establish a highly efficient pipeline for optimizing dynamic 3DGS from event streams and sparse RGB images.

\subsection{Event-based Neural Reconstruction}
Traditional RGB cameras suffer from motion blurs when the moving speed of cameras is fast and cannot be used in extreme lighting conditions because of their low dynamic ranges. However, most existing 3D reconstruction methods fail to provide event-based solutions. To address this issue, several works \cite{klenk2023enerf, hwang2023evnerf} have been proposed to incorporate the linearized event generation and NeRF pipeline. However, these works fail to reconstruct clear edges and textures and suffer from soft fogs caused by continuous NeRF networks. PAEv3D \cite{wang2024physical} introduces motion priors into the NeRF pipeline, enhancing the quality of edge and texture reconstruction. EvGGS \cite{wangevggs} introduces the first optimization framework to combine multiple event-based vision tasks with 3DGS. Some event-based 3DGS works reconstruct high-fidelity \cite{zahid20253dgs} or pose-free 3D scenes\cite{huang2025inceventgs, chen2025usp}, but fail in dynamic scenes since these previous works all regard the scenario as static. Nevertheless, DE-NeRF \cite{ma2023deformable} is the first to introduce event streams to model dynamic radiance fields and achieve satisfactory results. As we mentioned above, the sparsity and discontinuity of event data could lead to blurs and soft fogs because NeRF encodes scenes as a continuous network. To reconstruct the dynamic scene more faithfully, we incorporate event-based data and 3DGS and utilize the event flow extracted from the event stream for optimization. We experimentally prove that our reconstruction quality outperforms existing event-based and image-based methods.

\section{Methodology}
\subsection{Preliminaries and Overview}
\subsubsection{\textbf{Dynamic 3D Gaussian Splatting}}
\label{sec3.1}
Standard 3DGS represents the scene using 3D Gaussian points, each of which is characterized by several trainable parameters including position ($\mu \in \mathcal{R}^3$), quaternion ($\mathbf{q} \in \mathcal{R}^4$), scale factor ($\mathbf{s}\in \mathcal{R}^3$), spherical harmonics coefficients ($\mathbf{h} \in \mathcal{R}^{3(k+1)^2}$) and opacity ($o \in [0,1]$). The quaternion defines a $3 \times 3$ rotation matrix ($\mathbf{R}$). The 3D covariance matrix can be obtained by $\Sigma = RSS^TR^T$. Given a certain camera pose, one can render the novel view by projecting these 3D Gaussians into a 2D plane by blending depth-ordered Gaussians overlapping that pixel.
\begin{equation}
    C = \sum_{i\in \mathcal{N}}c_i\alpha_i\prod_{j=1}^{i-1}(1-\alpha_j)
\end{equation}
where $c_i$ refers to the color of the Gaussian $i$. $\alpha_i$ denotes the soft occupation of Gaussian $i$ at the certain 2D location, which can be obtained by :
\begin{equation}
     \alpha_i(x) = o_iexp(-\frac{1}{2}(x-\mu_i)^T{\Sigma}_{i}^{T}(x-\mu_i))
     \label{eq: computing opacity}
\end{equation}
In Eq.~\ref{eq: computing opacity}, $\Sigma$ and $\mu$ specifically refer to their 2D-projected version while they keep their original 3D meanings in other equations. 
The reconstruction loss between renderings and groundtruth images is used to optimize all Gaussian parameters. 

The conventional 3DGS only focuses on static scene reconstruction. Due to its flexible and explicit features, it is easy to be extended to reconstruct 4D dynamic scenarios. The most intuitive way is to separately train multiple 3DGS in each timestep and then interpolate between these sets (dynamic 3DGS tracking). However, it falls short of continuous monocular captures within a temporal sequence and leads to excessive memory consumption. A more prevailing way in the community is to jointly learn a deformation field along with the canonical 3D Gaussians. The deformation field transforms each canonical Gaussian point's position, rotation, and scale parameters into the corresponding values at the target timestamp, shown in Eq.~\ref{eq: deformation field}.
\begin{equation}
    (\delta \mathbf{x}, \delta \mathbf{s}, \delta \mathbf{q}) = \mathcal{F}_\Phi(\gamma(\mathbf{x}), \gamma(\mathbf{t}))
    \label{eq: deformation field}
\end{equation}
where $\gamma$ denotes the positional encoding. 
The representative works are Deform-GS \cite{yang2023deformable} and 4D-GS \cite{wu20234d}, which can effectively learn 4D scene representation from monocular videos. The Deform-GS adopts a neural network to implement $\mathcal{F}$ while 4D-GS uses a triplane representation for that. In our work, we follow the Deform-GS to utilize a neural network to model the deformation field. 

\subsubsection{\textbf{The Work Principle of Event Camera}}
We introduce the work principle of the event camera. An event camera carries independent pixel sensors that operate continuously and generate "events" $e_k=(\mathbf{x}_k, t_k, p_k)$ whenever the logarithmic brightness at the pixel increases or decreases by a predefined threshold. Each event $e_k$ includes the pixel-time coordinates $(\mathbf{x}_k=(x_k, y_k), t_k)$ and its polarity $p_k=\{ +1, -1\}$. The most intuitive thought is that the events are triggered by the motion of the edge information in the scenes. The event trajectories are defined as the associations between events across different timestamps, which can be represented by the spatial-temporal event flow.

\subsubsection{\textbf{Overall Statement}}
\textbf{Problem Statement}.
The image-based dynamic 3DGS can be optimized from high frame-rate videos. However, it cannot be optimized from event camera data, i.e. the high temporal resolution event stream accompanied by sparse image sequences. Given a video captured by an event camera, including the event stream $\mathbf{[E_k]}_1^k$ and sparse RGB $[I_n]_1^N$, our goal is to reconstruct the high-quality dynamic scene from the two modalities. 

\noindent \textbf{Method Overview}
The primary reason why optimizing dynamic 3DGS from videos with low temporal resolution is difficult can be summarized as follows. Large deformation between two frames causes the solution ambiguity. For example, the number of paths to reach the same point in the second frame from the corresponding point in the first frame will be uncertain. In this case, the deformation field cannot be easily optimized. However, the inter-frame information provided by event cameras, i.e. the event trajectories, can be used to guide the training of the deformation field. Therefore, the key issues become 1. how to determine the event trajectories and 2. how to build motion correspondence between 2D event data and 3D Gaussian points. Since the motion of Gaussians in the same area can be considered similar, training for their scale and rotation parameters is not a major issue \cite{wangevggs}.

\begin{figure*}[t]
  \centering
  \includegraphics[width=\textwidth]{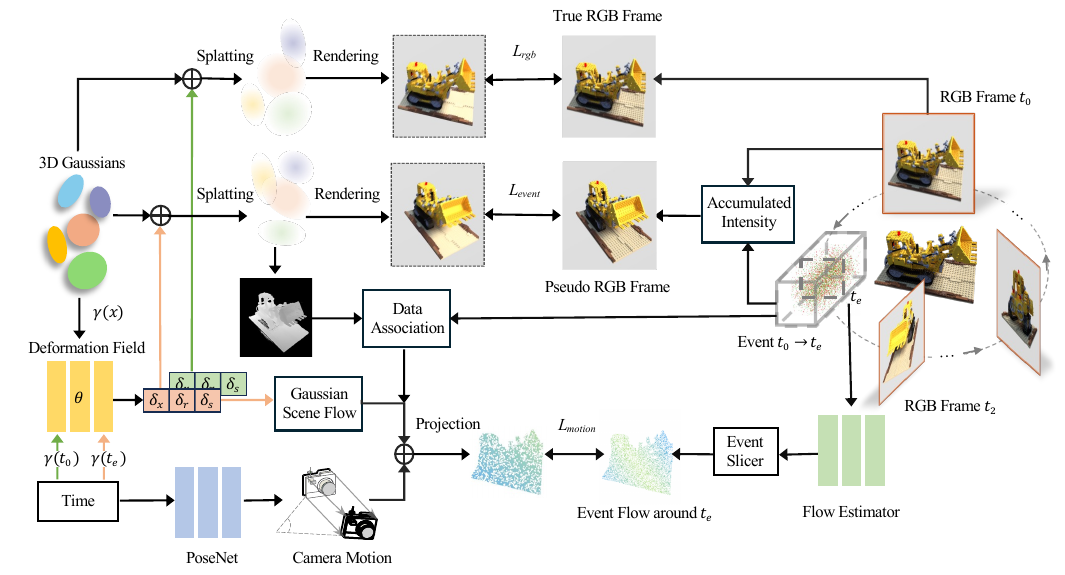}
  \caption{The overview of our proposed framework. Firstly, we fine-tune the optical flow estimator using LoCM in an unsupervised manner across the scene. Subsequently, we initialize canonical 3D Gaussians from sparse-view RGB sequences. Through Event-Gaussian Data Association, we establish correlations between 2D event data and 3D Gaussians via depth map registration, while continuously optimizing these associations during training. Our pipeline decomposes scene motion through event stream, enabling joint supervision of dynamic 3DGS using both event and RGB modalities.}
  \label{fig: main pipeline}
  \vspace{-2mm}
\end{figure*}

\subsection{\textbf{Contrast Maximization}}
In this section, we simply introduce the mathematical foundations of Contrast Maximization (CM) \cite{gallego2018unifying, Shiba_2022, shiba2022event, stoffregen2019event} as follows.
Assume a set of events $\mathbf{E}={e_k}_{k=1}^N$ are given, the goal of CM is to obtain the point trajectories on the image plane, which are described by the per-event optical flow. The image of warped events (IWE) can be obtained by aggregating events along candidate point trajectories to a reference timestamp\deleted{:}\added{.}   
\begin{equation}
    \mathbf{x}_k^{'} \doteq \mathbf{W}(\mathbf{x}_k, t_k;\mathbf{\theta}) =  \mathbf{x}_k-(t_k-t_{ref})\mathbf{v}_\mathbf{\theta}(\mathbf{x}_k,t_k)
    \label{eq: warp events}
\end{equation}
Here $\mathbf{W}$ is the warp function based on the event optical flow field $\mathbf{v}_\theta$, and $\mathbf{x}_k^{'}$ denotes the pixel coordinate of the warped event frame. If the input flow is correct, this reverses the motion in the events, and results in sharper event images. Thus the initial objective function is often defined to maximize the contrast of the IWE, given by the variance:
\begin{equation}
\begin{split}
Var(IWE(\mathbf{x};\mathbf{\theta})) &\doteq \frac{1}{|\Omega|}\int_\Omega(IWE(\mathbf{x};\mathbf{\theta})-\mu_I)^2 d\mathbf{x}\\
\mu_I &=\frac{1}{|\Omega|}\int_\Omega(IWE(\mathbf{x};\mathbf{\theta}))
\label{eq: contrast of IWE}
\end{split}
\end{equation}
The initial objective function measures the goodness of fit between the events and the candidate motion curves (warp). However, if we directly adopt the above IWE as the unsupervised loss, this might encourage all events to accumulate to several certain pixels. Inspired by \cite{Shiba_2022}, we adjust the objective by measuring event alignment using the magnitude of the IWE gradient to overcome the issue. Finally, we could define the refined objective function using the squared gradient magnitude of the IWE:
\begin{equation}
    Var(\theta;t_{ref}) \doteq \frac{1}{|\Omega|}\int_\Omega||\nabla IWE(\textbf{x};t_{ref}, \theta)||^2d\textbf{x}
    \label{eq: squared gradient of IWE}
\end{equation}

\subsection{Extracting Motion Priors from Events Using LoCM}
\label{sec: extracting priors}
Firstly, We introduce how to extract event trajectories using the LoCM (Low-Rank Adaptation+Contrast Maximization) framework in an unsupervised manner in this section.
The event flow estimators\cite{Zhu_2018, zhu2019unsupervised, gehrig2021raft, hagenaars2021self} can predict event flow by receiving raw event data as input. These models have been trained on various large datasets with per-event groundtruth. However, due to the domain gap between data distribution of the training dataset and the scene we need to reconstruct, the performance of the estimators would degrade when they are applied to unobserved scenarios, especially those with different motion patterns. A common way to use these pretrained event flow estimators is to finetune them on unseen scenes, but full-parameter finetuning struggles with two issues, 1. the lack of groundtruth of event flow and 2. overfitting on small dataset overturns the original motion priors in these models, which are learned from large datasets 

Inspired by NeuMA\cite{cao2024neuma}, we consider the prior knowledge of the estimator exists in its parameter space. Since we want the model to achieve better performance in new scenarios while preserving its prior knowledge, it is necessary to align the model to the out-of-distribution domain while maintaining its initial parameters.Therefore, to address the two challenges, we present to combine low-rank adaptation \cite{hu2021lora} with Contrast Maximization framework \cite{shiba2022event, gallego2018unifying} to finetune the estimator in an unsupervised manner. The scene-specific finetune can be expressed as:
\begin{equation}
    W_\theta := W_0 + \Delta W_\theta
    \label{eq: flow net finetune}
\end{equation}
where $W_0$ is the frozen initial parameter of the event flow estimator and we implement the trainable bypass $\Delta W_\theta$ with LoRA at $\Delta W_\theta=BA$. We adopt a random Gaussian initialization for $A$ and zero for $B$, so $\Delta W_\theta=BA$ is zero at the initialization of training. Zero initialization can restrict the correctness not deviating so much from the original value. Moreover, using LoRA can help the correction not to overturn the motion priors. 

In addition, due to the lack of groundtruth of event flow, we propose to finetune the event flow estimator in an unsupervised manner. The CM framework  is an effective proxy for extracting motion information from raw event data. The idea behind this optimization framework is that motion models can wrap events triggered by the same portion of a moving edge to produce a sharp event-accumulated image. 
Instead of directly optimizing the IWE, which is sensitive to the arrangement (i.e., permutation) of the IWE pixel values, the optimization target can be regarded as the variance of the IWE. Moreover, To mitigate overfitting, we divide the image plane into a tile of non-overlapping patches and up/down-sampled to multi-scale branches \cite{Shiba_2022}. 

The objective function in Eq.\ref{eq: squared gradient of IWE} can then be iteratively optimized to obtain the optimal parameters $\theta$ for the event flow field. However, in this work, we use an event flow estimator to replace the warp function in Eq. \ref{eq: warp events}. Instead of using the warp function, we use the estimator to obtain the event flow for warping event images. This approach is designed because the pre-trained estimator has already learned prior knowledge about various motion patterns, such as simultaneous camera ego-motion and multi-object motions within the scene. This prior knowledge makes it easier to model unseen and complex motion patterns. In contrast, motion fields based on simple parameters struggle to distinguish between different motion patterns in the scene. From Eq.~\ref{eq: squared gradient of IWE}, we can derive the unsupervised learning loss as 
\begin{equation}
    L_{flow} = 1/Var(f_\theta(\mathbf{E}(t_1,t_{ref}))) + \lambda TV(\theta)
    \label{eq: optical flow loss}
\end{equation}
where $f_\theta$ is the flow prediction network, $\mathbf{E}(t_1,t_{ref})$ is the events between $t_1$ and $t_{ref}$. $\lambda$ denotes the weight to balance the CM unsupervised loss and the $TV$ regularization \cite{rudin1992nonlinear}. The $f_\theta$ predicts the event flow from $t_1$ to $t_{ref}$ of all events in $\mathbf{E}(t_1,t_{ref})$. Maximizing the contrast is equivalent to minimizing its reciprocal. We provide detailed architecture of the event flow estimator network in Sec.\ref{Sec.architecture} and implementation details of LoCM finetuning in Sec.\ref{sec:efficiency}.

By adopting such a LoCM framework, we can efficiently finetune the event flow estimator on the specific scene that we aim to reconstruct. The predicted event flow is, in the following, used to guide the optimization of the deformation field. 
\subsection{Event-Gaussian Data Association}
\label{sec: data association}
In this section, we introduce how to build the data association between 2D events and 3D Gaussians, which is fundamental for optimizing 3DGS with the help of event streams. As discussed in Sec.~\ref{sec3.1}, the core challenge is the uncertainty caused by the large motions between frames. In other words, the $\delta x$ in Eq.~\ref{eq: deformation field} is hard to optimize, which can be considered as the scene flow of each 3D Gaussian. To align the motion of events with the deformation of 3D Gaussians, we first establish the data associations between events and their corresponding 3D Gaussians. The differentiable renderer of 3DGS can smoothly produce the depth map for a given camera viewpoint at a certain timestamp because the $\alpha$-blending in rendering naturally deals with transparency and occlusion relationship. For a specific timestamp $t$, we produce its depth map at first. Since dynamic 3DGS experiences a warm-up stage (\ref{sec3.1}), the depth map can also be initialized well in a coarse manner. Then the 3D location to trigger each event can be found by unprojecting the depth value of each event to 3D space. The nearest 3D Gaussian to the unprojected 3D location is treated to be the most contributory to this event. 
In addition, we implement the binding correspondence in a soft manner for better robustness. 

In detail, each 3D Gaussian corresponds with its top-k nearest unprojected events (k=3 empirically) and generates a correspondence weight via the inverse distance weight function for them. Moreover, to avoid inaccuracy event-Gaussian binding caused by the coarse depth map, we firstly initialize the canonical 3D Gaussians set in a warm-up phase with 3500 iterations by employing the sparse RGB sequences to obtain an initialized coarse depth map to implement event-Gaussian data association. In the joint training phase, the depth map will be optimized, and we periodically update the event-gaussian bindings using optimized depth maps to better align the 2D event motion with the motion of 3D Gaussians. At present, the one-to-k Gaussian-to-Events data associations are established by using the geometry information obtained from 3DGS rasterization.

\subsection{Decomposed Motion Supervision}
However, the motion contributing to events contains not only the Gaussian deformation but also the camera ego-motion. Therefore, we decompose the motion signal into the Gaussian scene flow and the camera ego-motion. Unlike other NeRF-based approaches to interpolate camera trajectory or use turnable poses, we follow \cite{ma2023deformable} to use a PoseNet to generate a continuous pose function that maps time to the camera pose vector representation $(R, t)$. The PoseNet is proved to be very effective and a plug-play module in DE-NeRF \cite{ma2023deformable}. It is fed with the interpolated camera poses as inputs and outputs a corrected term to make it perfect. It is efficient to be trained because even though we directly adopt the interpolated camera poses, we can still obtain relatively not-bad results. 

In this context, given two nearby specific poses, we can easily derive their instantaneous translation and rotation velocities ($\mathbf{v}_c=[v_x, v_y, v_z]$ and $\mathbf{w}_c=[w_x, w_y, w_z]$) by assuming the camera is moving rigidly in a small time interval. The pixel-level image velocity caused by the camera ego-motion can be derived from the camera's rigid motion and its corresponding 3D location:
\begin{equation}
   \mathbf{F}_{ego}\!\! =\!\! \frac{1}{Z}\left (\!\!\!\begin{array}{ccc} -1\!\!\!\!\! & 0\!\!\!\!\! & x\!\!\! \\ 0\!\!\!\!\! &-1\!\!\!\!\! &y\!\!\! \end{array}\right)\mathbf{v}_c+\left (\!\!\!\begin{array}{ccc} xy\!\!\!\!\! &-1-x^2\!\!\!\!\! & y\!\!\! \\ 1+y^2\!\!\!\!\! &-xy\!\!\!\!\! &-x\!\!\! \end{array}\right) \mathbf{w}_c
    \label{eq: camera ego motion}
\end{equation}
The detailed step-by-step derivation of the above equation can be found in \cite{mitrokhin2019ev, zhu2019unsupervised, Zhu_2018}. The $\mathbf{v}_c$ and $\mathbf{w}_c$ are projected to the image plane given the inverse depth ($\frac{1}{Z}$ in Eq.~\ref{eq: camera ego motion}) rendered from the 3DGS rasterization. 
Likewise, the Gaussian scene flow between two selected timestamps can be derived by:
\begin{equation} 
\begin{aligned}
\mathbf{F}_{gs}^i=Proj(\mathbf{x_i}+\mathcal{F}_\Phi(\gamma(x_i), \gamma(t_1))[0]) \\
- Proj(\mathbf{x_i}+\mathcal{F}_\Phi(\gamma(x_i), \gamma(t_0))[0])
    \label{eq: gaussian scene flow}
\end{aligned}
\end{equation}
Here the $[0]$ refers to selecting the first output term of the deformation field $\mathcal{F}_\Phi$. The $Proj$ function denotes the projection of a 3D location to the 2D image plane by camera transformation matrix. $i$ refers the $i$-th Gaussian. Until now, we have established the decomposed motion correspondence from 3D to 2D. The combination of $\mathbf{F}_{gs}$ and $\mathbf{F}_{ego}$ can be optimized through the finetuned event flow estimator presented in Sec.\ref{sec: extracting priors}, which we will introduce in detail in the following section.  
\begin{figure*}[t!]
  \centering  \includegraphics[width=0.98\textwidth]{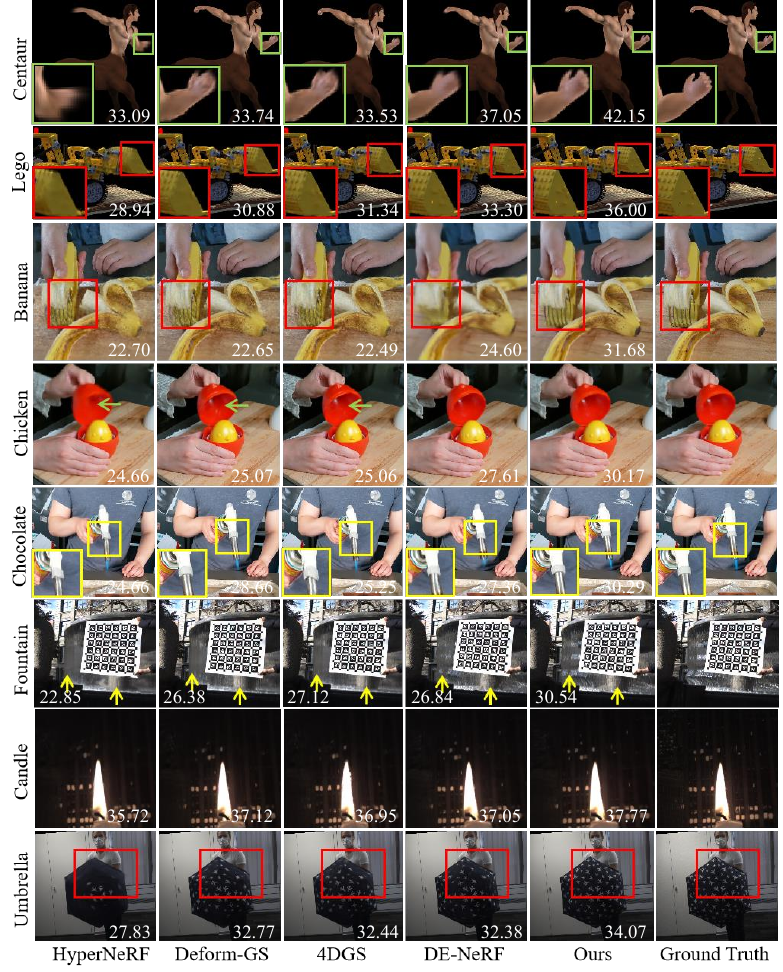} %
  \caption{\textbf{Qualitative Comparison on synthetic and real-world datasets.} Regions with notably different reconstruction qualities are highlighted with colored boxes and arrows.}
  \vspace{-7mm}
  \label{fig:comparison1}
\end{figure*}
\subsection{Training Paradigm}
\label{sec: training paradigm}

This section introduces the detailed training pipeline to optimize dynamic 3DGS by events and RGB modalities simultaneously. We first initialize the Gaussian point set in a warm-up phase with 3500 iterations by merely using the sparse RGBs. This is typically supervised by $L_{rgb}=||I_c-R(c)||_1^1$ where $c$ is the camera pose. In the second joint training phase, the $L_{rgb}$ remains unchanged. But we additionally integrate events into the training pipeline using our previously presented approach. At each iteration step, two consecutive images and the event between them are read out by the dataloader. Here we refer to the timestamp of the two images as $t_0$ and $t_2$. Next, we randomly select a timestamp between $t_0$ and $t_2$, which is referred to as $t_e$, for the subsequent event-part optimization. 

We leverage the events to optimize the information between two adjacent RGBs in a monocular video. In detail, at each optimization step, in addition to the RGB, we randomly select another timestamp around the timestamp of the selected RGB, we term it as $t_e$. Then we generate a pseudo image at the $t_e$ by using the events between the selected RGB and the $t_e$. The pseudo image can be considered a pseudo label for optimizing the 3DGS. 
The procedure for using the pseudo image to optimize 3DGS can be described as follows:
\begin{equation}
    L_{event}=\left\{
\begin{array}{lc}
||R(t_e)-I(t_0)e^{\Sigma_{e \in \Delta t}p_iC}||_{1}^{1}, & t_e \leq \frac{t_0+t_2}{2} \\
||R(t_e)-I(t_2)/e^{\Sigma_{e \in \Delta t}p_iC}||_{1}^{1}, & t_e > \frac{t_0+t_2}{2}
\end{array}
\right.
    \label{eq: event loss}
\end{equation}
$t_0$ and $t_2$ refer to the timestamps of the left and right RGB images. To mitigate the occlusion effect, we use the nearest RGB to estimate the color at $t_e$. 

Moreover, we adopt the flow estimator ($f_\theta(\mathbf{E};t_{ref})$ in Eq.~\ref{eq: optical flow loss}) to estimate the event trajectories, i.e. event optical flow, from $t_2$ to $t_0$. The $t_{ref}$ in this phase should be constantly set to the timestamp of the nearest RGB in Eq.~\ref{eq: event loss}. Here for the convenience of illustration, we assume $t_e < \frac{t_0+t_2}{2}$ so as $t_{ref}=t_0$. As we discussed in the previous section, the event flow is triggered by two decomposed motions. After applying the flow estimator, the optical flow of all events from $t_0$ to $t_2$ can be resolved. We then bind these $k$ events near $t_0$ to their corresponding 3D Gaussians based on the approach introduced in Sec.~\ref{sec: data association}. We use a small time interval to filter out candidate events $t_0 - \delta t \leq t \leq t_0 + \delta t$. Only considering the events whose timestamps are near $t_0$ is necessary because the same object might trigger independent events at different timestamps. Each Gaussian corresponds to at most $k=3$ events and at least 0 events. Next, we jointly train the ego-motion PoseNet and Gaussian deformation by :
\begin{equation}
    L_{motion} = \sum_{i=1}^N (\sum_{k \in \mathcal{B}(i)} f_\theta(\mathbf{E}(t_0,t_2)) - (\mathbf{F}_{ego}+\mathbf{F}_{gs}^i))
    \label{eq: motion loss}
\end{equation}
where $i$ denotes the $i$-th Gaussians with at least one associated event. $\mathcal{B}(i)$ illustrates the bond operation to establish event-Gaussian correspondence. Other symbols remain the same in the previous context. We set $k=3$ experimentally.
After the warm-up phase, the total loss is:
\begin{equation}
    L_{total} =L_{rgb}+\gamma_1 L_{event}+\gamma_2(iter) L_{motion}
    \label{eq: total loss}
\end{equation}
$\gamma_1$ and $\gamma_2$ are hyperparameters to balance the magnitude of these items. We set $\gamma_1=1$ and $\gamma_2=1-e^{-\frac{1}{4000} iter}$ that is a function of training iterations. At the early training stage, the deformation field cannot predict accurate Gaussian scene flow. We use an annealing function to weigh this term.

\begin{figure*}[t!]
    \centering
    \includegraphics[width=1.0\textwidth]{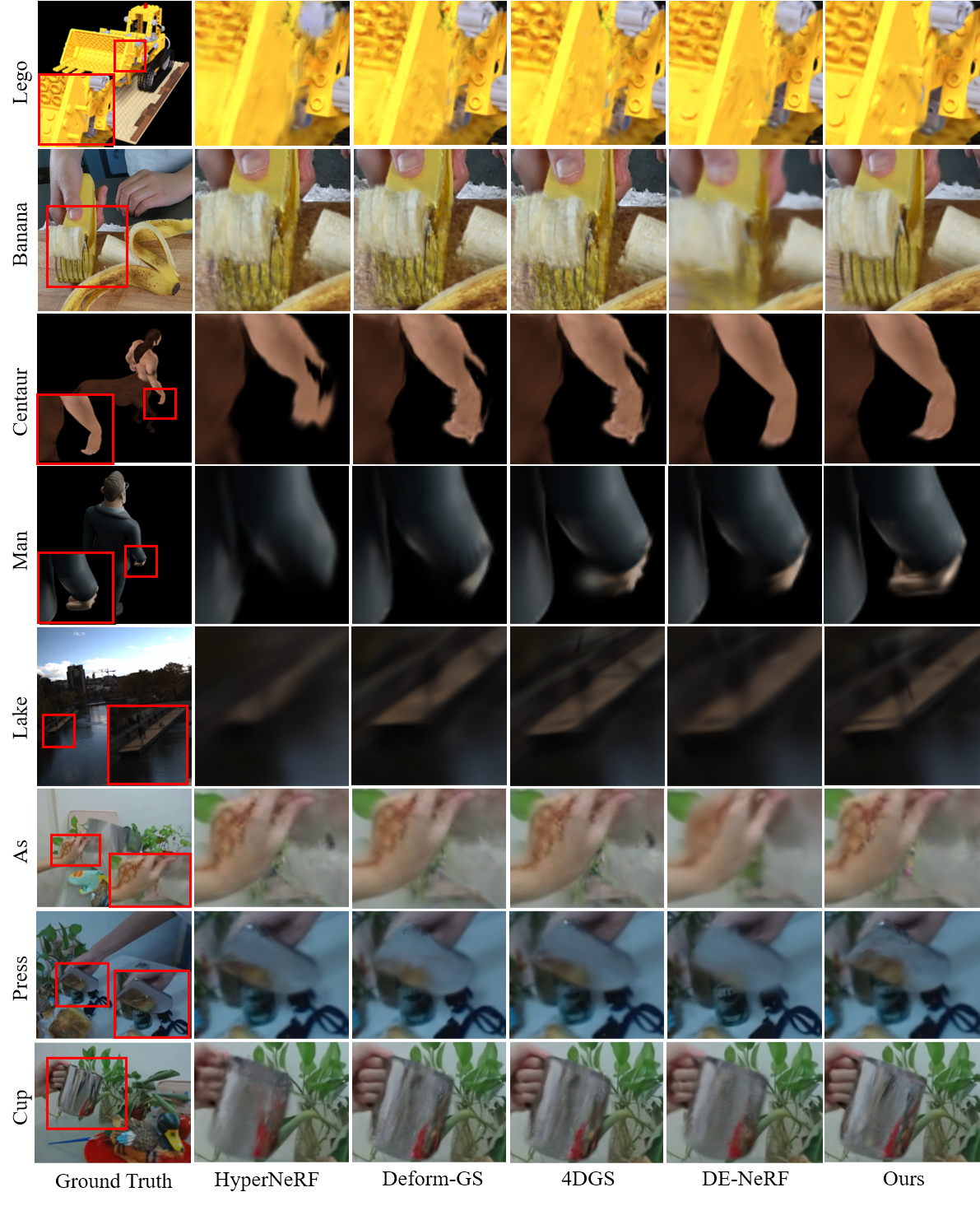}
    \caption{\textbf{Comparison visualization based on enlarged images.} The Cup, Press and As scenarios are from the NeRF-DS dataset, and Lake is another scene from the HSERGB dataset. Lego and Centaur are our novel synthetic datasets. Banana is from the HyperNeRF dataset.}
    \label{fig:appendix}
    \vspace{-3mm}
\end{figure*}\section{Experiments}
\label{sec: experiments}
In this section, we thoroughly validate the effectiveness of the proposed approach both on the synthetic and real-world datasets and ablate the components constituents contained in this approach. 




\begin{table*}[!t]
\centering
\resizebox{\textwidth}{!}{
\begin{tabular}{lccccccccc} 
\hline
\multicolumn{1}{c}{} & \multicolumn{3}{c}{\textbf{Centaur}}   & \multicolumn{3}{c}{\textbf{Lego}}                        & \multicolumn{3}{c}{\textbf{Man}}                                                                                                              \\
Methods& \multicolumn{1}{c}{PSNR↑}& \multicolumn{1}{c}{SSIM↑}& \multicolumn{1}{c}{LPIPS↓} & \multicolumn{1}{c}{PSNR↑}& \multicolumn{1}{c}{SSIM↑} & \multicolumn{1}{c}{LPIPS↓} & \multicolumn{1}{c}{PSNR↑}& \multicolumn{1}{c}{SSIM↑} &\multicolumn{1}{c}{LPIPS↓}\\ 
\hline
HyperNeRF & 33.09 & 0.9814 & 0.0256 & 28.94 & 0.9545 & 0.0487 & 33.25 & 0.9758 & 0.0281\\
Deform-GS & 33.74 & 0.9829 & 0.0187 & 30.88  & 0.9646  & 0.0350 & 34.98 & 0.9805  & 0.0221 \\
4DGS   & 33.53 & 0.9819 & 0.0194 & 31.34 & 0.9662 & 0.0322 & \underline{37.60} & \underline{0.9817} & \underline{0.0184}  \\
DE-NeRF & \underline{37.35} & \underline{0.9895} & \underline{0.0149} & \underline{33.30} & \underline{0.9780} & \underline{0.0261} & 35.87 & 0.9815 & 0.0194  \\
Ours & \textbf{42.15}  & \textbf{0.9971}  & \textbf{0.0076}  & \textbf{36.00}  & \textbf{0.9916}  & \textbf{0.0120}  & \textbf{39.37}  & \textbf{0.9856}  & \textbf{0.0149}   \\
\hline
\end{tabular}}
\caption{Quantitative comparisons of baselines and ours on synthetic scenes.}
\label{table:metric_syn}
\vspace{-4mm}
\end{table*}

\begin{table*}[!t]
\centering
\resizebox{\textwidth}{!}{
\begin{tabular}{lccccccccc} 
\hline
& \multicolumn{3}{c}{\textbf{Candle}} & \multicolumn{3}{c}{\textbf{Umbrella}} & \multicolumn{3}{c}{\textbf{Fountain}} \\
Methods   & \multicolumn{1}{c}{PSNR↑} & \multicolumn{1}{c}{SSIM↑} & \multicolumn{1}{c}{LPIPS↓}   & \multicolumn{1}{c}{PSNR↑} & \multicolumn{1}{c}{SSIM↑} & \multicolumn{1}{c}{LPIPS↓} &\multicolumn{1}{c}{PSNR↑} & \multicolumn{1}{c}{SSIM↑} & \multicolumn{1}{c}{LPIPS↓}\\ 
\hline
HyperNeRF & 35.72 & 0.9489 & 0.2512 & 27.83 & 0.8379 & 0.2687 & 22.85 & 0.8036 & 0.5013\\
Deform-GS & \underline{37.12} & \textbf{0.9579}  & \underline{0.2266}  & \underline{32.77} & \textbf{0.8611} & \underline{0.1557} & 26.38 & 0.8516 & 0.4688 \\
4DGS & 36.95 & \underline{0.9569} & 0.2299 & 32.44 & 0.8598  & 0.1617  & \underline{27.12} & \underline{0.8717} & 0.4547\\
DE-NeRF & 37.05 & 0.9546 & 0.2405 & 32.38 & \underline{0.8609} & 0.1591 & 26.84 & 0.8523 & \underline{0.4345}\\
Ours & \textbf{37.77}  & 0.9450 & \textbf{0.1132} & \textbf{34.07}  & 0.8513 & \textbf{0.1210} & \textbf{30.54} & \textbf{0.9125} & \textbf{0.0277}\\
\hline
& \multicolumn{3}{c}{{\textbf{Banana}}}  & \multicolumn{3}{c}{{\textbf{Chicken}}}  & \multicolumn{3}{c}{{\textbf{Chocolate}}}\\
Methods   & \multicolumn{1}{c}{PSNR↑} & \multicolumn{1}{c}{SSIM↑} & \multicolumn{1}{c}{LPIPS↓}   & \multicolumn{1}{c}{PSNR↑} & \multicolumn{1}{c}{SSIM↑} & \multicolumn{1}{c}{LPIPS↓} &\multicolumn{1}{c}{PSNR↑} & \multicolumn{1}{c}{SSIM↑} & \multicolumn{1}{c}{LPIPS↓}\\ 
\hline
HyperNeRF & 22.70 & 0.5167 & 0.3441 & 24.66 & 0.7589 & 0.2990 & 24.66 & 0.8452 & 0.1697\\
Deform-GS & 22.65 & 0.5067 & \underline{0.3294} & 25.07 & 0.7621 & 0.1959 & \underline{28.66}  & \underline{0.9041}  & \textbf{0.1218}   \\
4DGS & 22.49 & 0.4958 & 0.3353 & 25.06 & 0.7613 & \underline{0.1936}  & 25.25 & 0.8470 & 0.1466  \\
DE-NeRF  & \underline{24.60} & \underline{0.6206} & 0.4636 & \underline{27.61} & \underline{0.8323} & 0.2344 & 27.36 & 0.8916 & 0.1689\\
Ours & \textbf{31.68} & \textbf{0.9586} & \textbf{0.0950} & \textbf{30.17} & \textbf{0.9371} & \textbf{0.1526}  & \textbf{30.29}  & \textbf{0.9460}  & \underline{0.1240} \\
\hline
\end{tabular}}
\caption{Quantitative comparisons of baselines and ours on real-world datasets.}
\label{table:metric_real}
\end{table*}

\begin{table*}
\centering
\resizebox{\textwidth}{!}{
\begin{tabular}{lllllllccc}
\hline
& \multicolumn{3}{c}{\textbf{Cup}} & \multicolumn{3}{c}{\textbf{As}} & \multicolumn{3}{c}{\textbf{Press}} \\
Methods & \multicolumn{1}{c}{PSNR↑} & \multicolumn{1}{c}{SSIM↑} & \multicolumn{1}{c}{LPIPS↓} & \multicolumn{1}{c}{PSNR↑} & \multicolumn{1}{c}{SSIM↑} & \multicolumn{1}{c}{LPIPS↓} & PSNR↑ & SSIM↑ & LPIPS↓ \\
\hline
HyperNeRF& 23.55 & 0.8529 & 0.1986 & 26.69 & 0.9175 & \textbf{0.1573} & 25.38 & 0.8549 & 0.1978 \\
Deform-GS& 23.87 & 0.8591 & 0.1807 & \underline{26.75} & \underline{0.9189} & 0.1590 & 25.72 & \textbf{0.8735} & 0.2022\\
4DGS & 23.81 & 0.8601 & 0.1794 & 26.59 & 0.9165 & \underline{0.1580} & \underline{25.75} & 0.8655 & \underline{0.1917} \\
DE-NeRF & \textbf{24.25} & \underline{0.8708} & \underline{0.1791} & 26.24 & 0.9065 & 0.1962 & 24.51 & 0.8448 & 0.2382\\
Ours & \underline{24.10} & \textbf{0.8740} & \textbf{0.1728} & \textbf{27.18} & \textbf{0.9224} & 0.1581 & \textbf{25.82} & \underline{0.8691} & \textbf{0.1891} \\
\hline
\end{tabular}}
\caption{Quantitative comparisons on NeRF-DS dataset. We also compare our method against previous methods on three real-world scenes from the NeRF-DS dataset.}
\label{table:appendix}
\end{table*}

\textbf{Synthetic Dataset.} On account of the absence of relevant synthetic event-based monocular 4D reconstruction dataset. We establish a novel one with three scenarios. They carry varying degrees and types of large-scale motions. We adopted Blender to produce video camera poses and simulated event streams using V2E \cite{hu2021v2e}. 
\begin{figure}[!hb]
    \centering
    \includegraphics[width=0.5\textwidth]{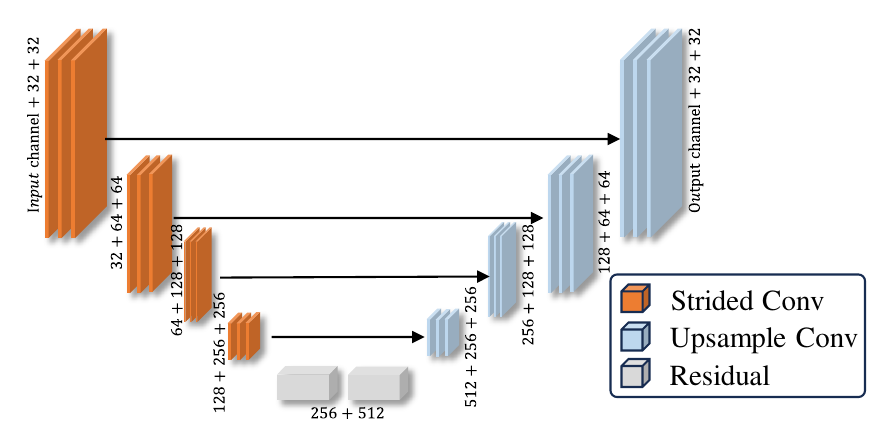}
    \caption{\textbf{The architecture of the event-based optical flow estimator}}
    \label{fig:evflownet}
\end{figure}

\textbf{Realistic Dataset.} We evaluate our methods and other counterparts on three real-world datasets. The first is the High-Speed Events and RGB dataset (HS-ERGB) \cite{Tulyakov21CVPR} which includes challenging and moving fastly dynamic scenes captured by a realistic event camera. We select three challenging scenes in the dataset, i.e. \textbf{Umbrella}, a rapid rotating scene, \textbf{Candle}, an environment with rapid jitter, and \textbf{Fountain} a super-fast liquid scenario. The dataset provides complete and realistic event streams, RGB sequences as well as camera parameters. The other two real-world datasets that we use in this work, the HyperNeRF dataset \cite{park2021hypernerf} and the NeRF-DS dataset \cite{yan2023nerfds}, are commonly used in purely image-based NeRF or 3DGS research. They only provide RGB videos but unfortunately without corresponding real event streams. Using simulators for event generation makes it difficult to replicate the various noises found in events captured by real event cameras. Therefore, we use a real event camera, DVXplorer \cite{Inivation} to convert the two datasets into their event-based versions to obtain realistic images and event data pairs for the convenience of comparison with conventional image-based methods. The details of the data collection pipeline and collection system setup are illustrated in \ref{sec.data collection} 

\textbf{Blurry Dataset.} To evaluate the robustness to motion blur, we further conduct an experiment on scenes with motion blur. For the blurry image dataset, we adopt the blurry image sequences from the BARD Dataset \cite{lu2025bard}. This dataset includes various types of complex motions in indoor scenes, captured using a GoPro camera to record continuous monocular motion RGB sequences. For each scene, we separately calibrate the images sequences using COLMAP to obtain their intrinsic and extrinsic parameters and then create blurred images from clean scene images. Utilizing the pre-trained Video Frame Interpolation (VFI) model, RIFE\cite{huang2022rife}, we interpolate the frame rate of the original video to 8 times. These interpolated sharp frames are then averaged to simulate the blur formation process, resulting in blurred images. We then generate the synthetic event stream from the image sequences using the V2E event converter, while reserving the sharp images as our test set, and evaluate the performance of our framework. We evaluated six scenes: \textbf{kitchen, micro-lab, toycar, card, cube-desk, and windmill}. Each of which features complex environment settings, motion-blurred images, and different types of complex dynamic motions. We select the Deform-GS and 4DGS as the baseline, utilizing the blurry image for training and test the robustness to motion blur.


\subsection{Results and Comparisons}
In our experiments, we compare our method with four baselines, HyperNeRF \cite{park2021hypernerf}, 4DGS \cite{wu20234d}, Deformable GS (Deform-GS) \cite{yang2023deformable}, and DE-NeRF \cite{ma2023deformable}. The first three methods are currently prevailing image-based dynamic reconstruction approaches while the last one is the SOTA event-based dynamic NeRF method. We perform both quantitative and qualitative evaluations for a comprehensive and convincing comparison.

\textbf{Quantitative Comparison.}
Firstly, we quantitatively evaluate these methods on scene reconstruction quality metrics including PSNR, SSIM, and LPIPS. We employ the VGG network for LPIPS evaluation. As shown in Table \ref{table:metric_syn} and Table \ref{table:metric_real}, we compute the mean values for all metrics across scenes from both synthetic and realistic datasets. We use two symbols to indicate the top two performing methods in nine scenarios, with \textbf{bold}, and \underline{underline} representing the best and second best respectively. The results demonstrate that our proposed approach exhibits superior performance across all scenarios, proving the effectiveness of our framework. 

We also quantitatively evaluate our method and four baselines on the NeRF-DS dataset and conduct the evaluation experiment with metrics such as PSNR, SSIM, and LPIPS. The results are presented in Table.\ref{table:appendix} and indicate that our method achieves outstanding performance in all metrics and most of the scenes within the dataset. Cells are marked as \textbf{bold}, and \underline{underline}, representing the best, and second best respectively.

\textbf{Qualitative Comparison.}
We also provide qualitative results and comparisons for a better visual assessment. We visualize the results in Fig.~\ref{fig:comparison1}, and it can be observed that our method recovers more detailed information when synthesizing images from novel viewpoints. Our method reconstructs more delicate object contours and textures, especially in the regions annotated with boxes. Our approach surpasses existing methods by a great margin, demonstrating the remarkable capability of restoring the contents and details of the given scene over time. The effectiveness of our method benefits from the rich information derived from both RGB image modality and event modality.

As shown in Fig.\ref{fig:appendix}, We also provide additional qualitative results on novel scenes and previously evaluated scenes from new viewpoints. Certain regions in the images are magnified to compare the recovered details and demonstrate the differences in reconstruction quality between our method and other baselines. 

\begin{figure*}[h]
    \centering
    \centering
    \includegraphics[width=0.98\textwidth]{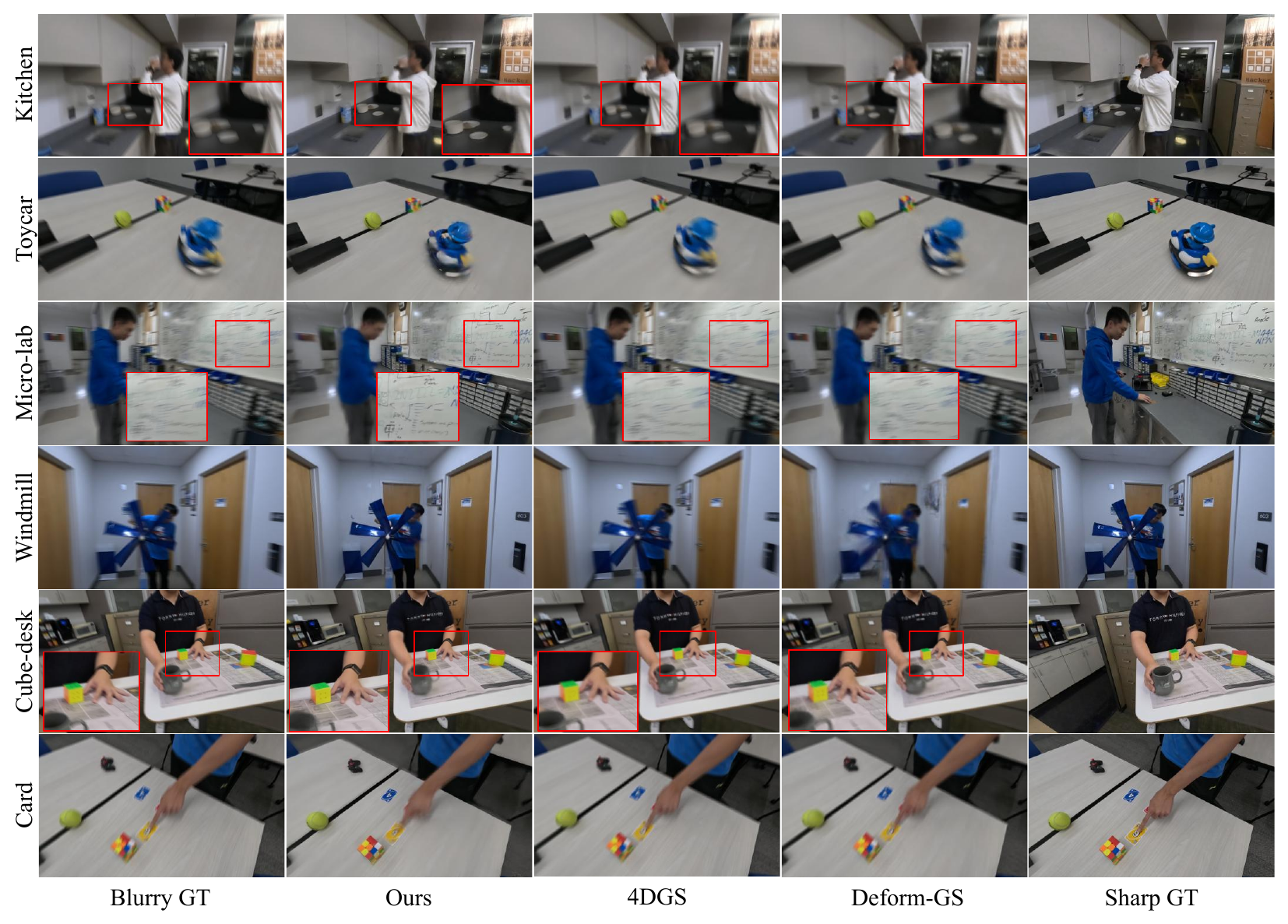}
    \caption{The qualitative results for robustness against motion blur.}
    \label{fig:blur}
\end{figure*}

\begin{table*}[ht]
  \centering  
  \resizebox{\textwidth}{!}{
  \begin{tabular}{lccccccccc}
    \toprule
    & \multicolumn{3}{c}{\textbf{Windmill}}
    & \multicolumn{3}{c}{\textbf{Cube‑desk}}
    & \multicolumn{3}{c}{\textbf{Card}} \\
    Methods& SSIM$\uparrow$ & PSNR$\uparrow$ & LPIPS$\downarrow$
    & SSIM$\uparrow$ & PSNR$\uparrow$ & LPIPS$\downarrow$
    & SSIM$\uparrow$ & PSNR$\uparrow$ & LPIPS$\downarrow$ \\
    \midrule
    Df‑GS
      & 0.8231 & 22.35 & 0.3027
      & 0.7833 & 22.80 & 0.2516
      & 0.8221 & 24.21 & 0.3601 \\
    4DGS
      & \underline{0.8317} & \underline{23.10} & \underline{0.2168}
      & \underline{0.7855} & \underline{22.91} & \underline{0.2354}
      & \underline{0.8234} & \underline{24.28} & \underline{0.3345} \\
    Ours
      & \textbf{0.9271} & \textbf{27.17} & \textbf{0.1791}
      & \textbf{0.9390} & \textbf{30.68} & \textbf{0.1141}
      & \textbf{0.8910} & \textbf{30.33} & \textbf{0.2953} \\
    \midrule
    & \multicolumn{3}{c}{\textbf{Kitchen}}
    & \multicolumn{3}{c}{\textbf{Toycar}}
    & \multicolumn{3}{c}{\textbf{Micro‑lab}} \\
    Methods& SSIM$\uparrow$ & PSNR$\uparrow$ & LPIPS$\downarrow$
    & SSIM$\uparrow$ & PSNR$\uparrow$ & LPIPS$\downarrow$
    & SSIM$\uparrow$ & PSNR$\uparrow$ & LPIPS$\downarrow$ \\
    \midrule
    Df‑GS
      & 0.7651 & 20.53 & 0.3464
      & 0.8581 & 23.74 & 0.3013
      & 0.6502 & 20.17 & 0.4310 \\
    4DGS
      & \underline{0.7661} & \underline{20.56} & \underline{0.3637}
      & \underline{0.8632} & \underline{24.24} & \underline{0.2707}
      & \underline{0.6521} & \underline{20.20} & \underline{0.4126} \\
    Ours
      & \textbf{0.9092} & \textbf{26.51} & \textbf{0.1834}
      & \textbf{0.9143} & \textbf{27.45} & \textbf{0.2467}
      & \textbf{0.8808} & \textbf{26.60} & \textbf{0.2097} \\
    \bottomrule
  \end{tabular}}
  \caption{Quantitative comparisons on blurry scenarios}
  \label{tab:blur}
\end{table*}

\subsection{Robustness to Moiton Blur}
We provide qualitative and quantitative comparisons of experiments on blurry image scenarios to show the robustness to motion blur of our proposed method and other baselines. The visualization results of the supplementary experiment are shown in Figure.\ref{fig:blur} and Table.\ref{tab:blur}. It can be observed that although deblur is not the main goal of our proposed method, we successfully restore more sharp details from the blurry images than other image-based baselines, which demonstrates our method’s robustness to motion blur.

\subsection{Ablation Study}
We report the contributions of each part of the training configurations and components of the proposed approach in this section. As shown in Table \ref{table:ablation},  "only $L_{rgb}$" denotes that our method degrades to the conventional deformable GS and only uses sparse RGBs to train the model. "w/o $L_{event}$" means $\gamma_1=0$ in Eq.~\ref{eq: total loss}, while "w/o $L_{motion}$ refers that $\gamma_2$ is constantly set to 0 in Eq.~\ref{eq: total loss}. "w/o Flow Ft" represents that we directly adopt the original weights that are open-source with its code of EvFlowNet to estimate event flow without using Eq.~\ref{eq: optical flow loss} to finetune it. "Full Ft" refers to that we normally fine-tune all parameters in the EvFlowNet without using the LoCM framework.  "w/o PoseNet" means we use pose interpolation between frames instead of the PoseNet to get the event pose, even though this has already been proved to be effective by \cite{ma2023deformable}. We experimentally demonstrate that the performance of the estimator degrates on unseen scenes, and our finetuning strategy effectively improves its performance. We validate all the components on the three scenes by comparing their respective PSNR, SSIM, and LPIPS. The results illustrate that all the above modules contribute differently to learning dynamic Gaussian fields from event data. Among these, the motion prior has a greater impact on the results, while the color prior and pose have less influence.
\begin{figure*}[h!]
    \centering
    \includegraphics[width=\linewidth]{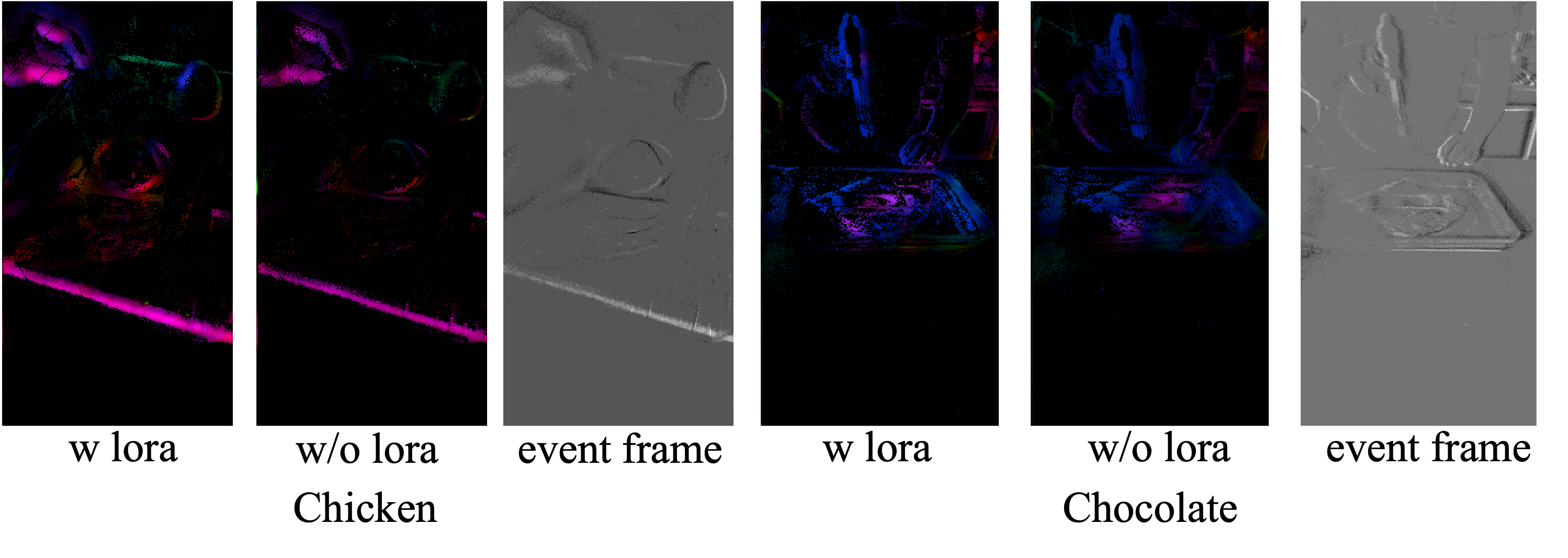}
    \caption{Qualitative examples of event flow prediction with or without LoCM fine-tuning. }
    
    \label{fig: flow map}
\end{figure*}
Moreover, we indicate two visual examples to show the superiority of the LoCM fine-tuning strategy in Fig.~\ref{fig: flow map}. It is observed that the predicted flows fine-tuned by LoCM ("w lora" in the figure) are sharper, and can better distinguish the object boundaries. In addition, they exhibit strong contrast between objects. Notably, quantitative results are not reported because our dataset, as we stated previously, is recorded by a realistic DVXplore, thereby no groundtruth of event-level flows. More comparison and analysis can be found in the following sections.

\begin{table}
\setlength\tabcolsep{1pt}
\centering
{
\begin{tabular}{ccccccc} 
\hline
& \multicolumn{3}{c}{\textbf{Lego}}   & \multicolumn{3}{c}{\textbf{Chocolate}}  \\
& PSNR↑   & SSIM↑   & LPIPS↓  & PSNR↑   & SSIM↑ & LPIPS↓ \\ 
\hline
only $L_{rgb}$  & 30.80    & 0.964   & 0.035 & 28.66  & 0.904   & 0.122  \\
w/o $L_{event}$  & 31.62   & 0.966  & 0.031 & 28.92 & 0.917 & 0.133  \\
w/o $L_{motion}$  & 31.24 & 0.968 & 0.031 & 25.65 & 0.861 & 0.183\\
w/o Flow Ft & 32.47 & 0.973 & 0.025 & 27.31 & 0.892 & 0.174  \\
Full Ft & 34.61 & 0.981 & 0.026 & 29.11 & 0.915 & 0.168  \\
w/o PoseNet  & 34.15 & 0.977 & 0.022 & 28.62 & 0.906 & 0.157  \\
unaltered & \textbf{34.89}  & \textbf{0.982}  & \textbf{0.021}  & \textbf{30.29}  & \textbf{0.946}  & \underline{0.124}   \\
\hline
\end{tabular}}
\caption{Ablations studies of different components in the proposed framework.}
\label{table:ablation}
\vspace{-5mm}
\end{table}

\begin{figure*}[ht]
    \centering
    \centering
    \includegraphics[width=0.98\textwidth]{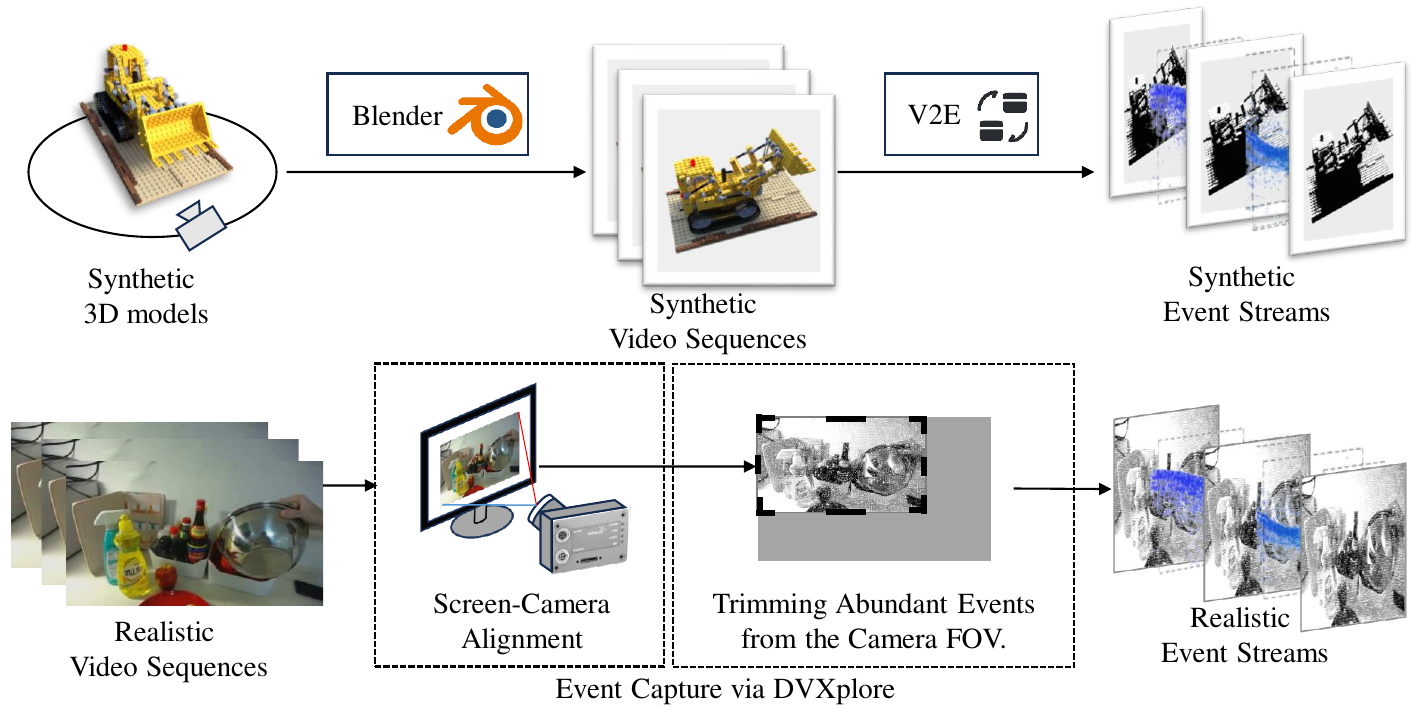}
    \caption{Data collection system. We leverage the realistic DVXplorer event camera and a high frame rate screen to convert the HyperNeRF and NeRF-DS datasets into their event-based version. }
    \label{fig:data collection}
\end{figure*}

\subsection{Depth Visualization}
\begin{figure}[b!]
    \centering
    \includegraphics[width=0.48\textwidth]{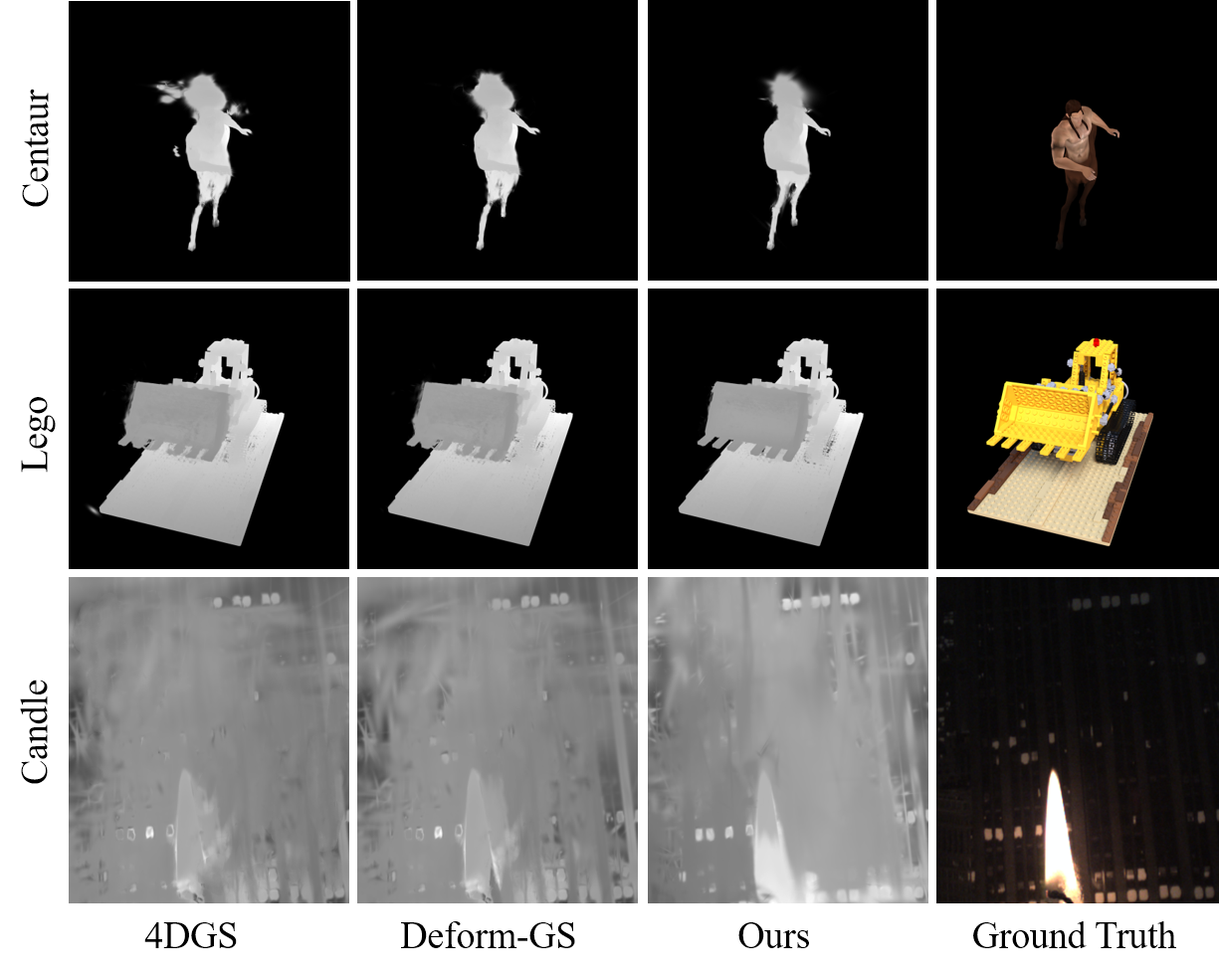}
    \caption{Depth visualization. We compare our method against 4DGS \cite{wu20234d} and Deform-GS \cite{yang2023deformable} on both synthetic and real-world scenes. These scenes are Centaur, Lego, and Candle from the top down.}
    \label{fig:depth visualization}
\end{figure}
As illustrated in Fig.\ref{fig:depth visualization}, we visualize the depth map of test scenes. Our proposed method yields substantially more accurate depth maps than other baselines, highlighting its superior geometric reconstruction capabilities. This underscores the method's efficacy across both synthetic and real-world datasets.

 \section{Implementation and Discussion}
\subsection{Experimental Settings}
We adopt the EvFlowNet \cite{Zhu_2018} as the $f_\theta$ in Eq.~\ref{eq: optical flow loss}. We first load its original network parameters then we use Eq.~\ref{eq: flow net finetune} and Eq.~\ref{eq: optical flow loss} to finetune it to fit a single scene event stream. We use Adam to optimize the EvFlowNet with an exponential decay learning rate from 0.0005 to 0.0001. We update the event-Gaussian binding every 500 iterations.

Moreover, for the selection of the contrast threshold, since all datasets were converted from RGB videos using a real event camera with known C, we faithfully adopted the hardware-calibrated Contrast Threshold C. In our experiments, the C is set to 0.1.

\subsection{Event Representation}
In this section, the representation of asynchronous raw event streams and how they are preprocessed as input for the optical flow estimator will be introduced in detail. Events ($e_k=(\textbf{u}_k,t_k,p_k)$) at pixel $\mathbf{u}_k=(u, v)$ and timestamp $t_k$ are triggered and output asynchronously, and the illumination change of the pixel can be represented using the polarity $(p \in \{+1, -1\})$. Thus, an event triggered at timestamp $t_k$ can be written as :
\begin{equation}
    \Delta L_k(\textbf{u})=\sum_{e_i\in\Delta t_k}p_iC
\end{equation}
Where L is the logarithmic frame ($L(t)=log(I(t))$) and C denotes the event trigger threshold value. Therefore, given threshold value $C$ and time interval $\Delta t$, triggered events can be accumulated in $\Delta t$, and thus obtain the log illumination changes. Due to the sparsity of the event streams, it is necessary to convert the asynchronous event streams at the given time interval $\Delta t$ to a synchronous representation. Thus. the event streams are encoded as a spatial-temporal voxel grid. The duration $\Delta t$ is divided into $B$ temporal bins following
\begin{equation}
        E(u,v, t_n) = \sum_{i}p_i\max(0,1-|t_n-t_i^*|)
\end{equation}
where $t_i^*$ is the normalized timestamp determined by the number of bins by $\frac{B-1}{\Delta T}$. We set the temporal bins $B=5$.

\subsection{Data Collection System}
\label{sec.data collection}
\begin{figure}[b!]
    \centering
    \includegraphics[width=0.5\textwidth]{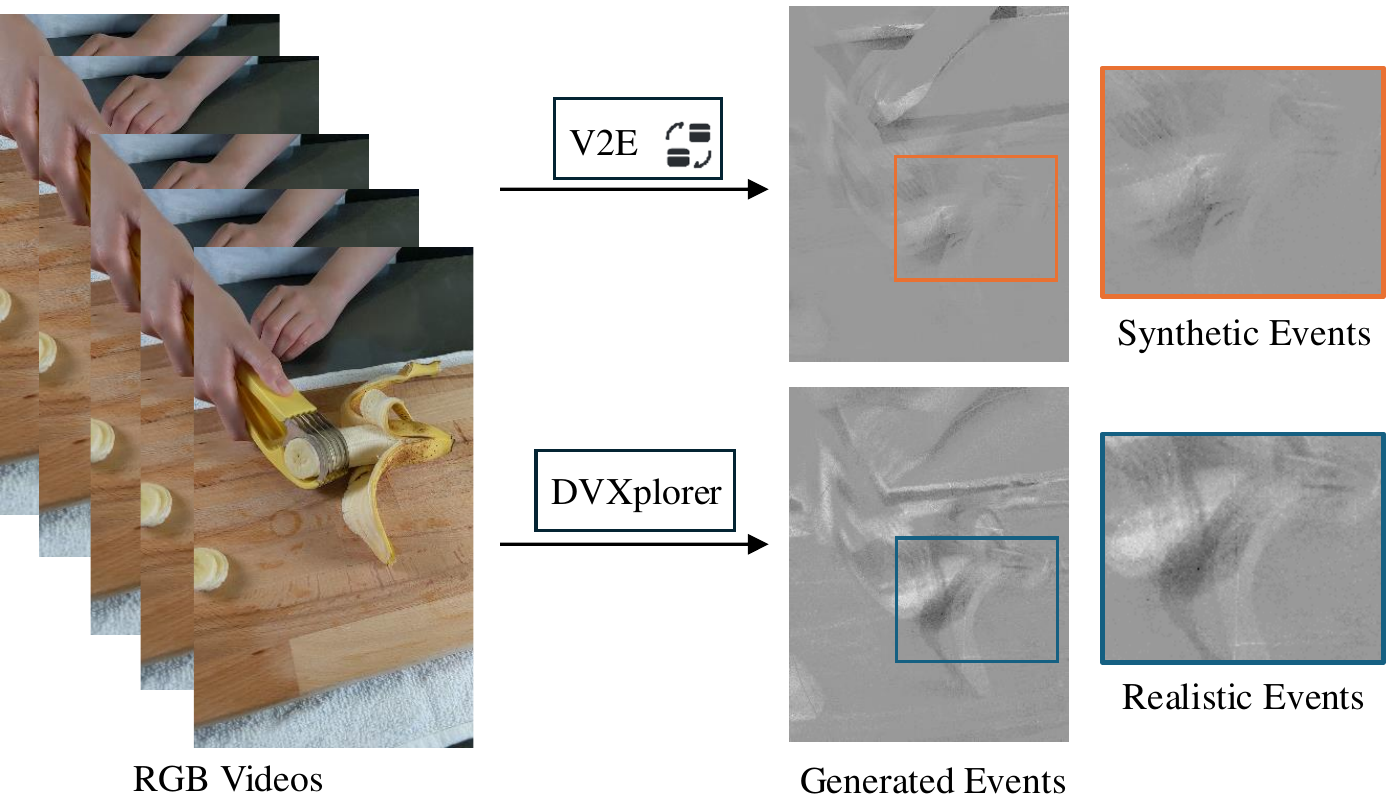}
    \caption{Visualization results of synthetic and realistic events of Banana scene}
    \label{fig:ev_vis}
    \vspace{-2mm}
\end{figure}
This section introduces our device setup for data collection. The data collection pipeline aims to produce corresponding event data for an RGB video. In this work, except for the HS-ERGB \cite{Tulyakov21CVPR}, which contains complete and realistic event streams of high-speed scenes, the other two datasets we utilize, the Hypernerf dataset \cite{park2021hypernerf} and the NeRF-DS dataset \cite{yan2023nerfds}, are image-based datasets, which only contain RGB video sequences. An intuitive method is to utilize an event camera simulator to convert the video into an event stream, as we have done in synthetic data. However, the synthetic event stream is not realistic enough. The main reason is that the event captured by a realistic event camera contains lots of irregular noises due to fluctuations of electronic components such as hot noises which do not exist in synthetic data. Even though one can add handcrafted noises to the simulated event stream, such as Gaussian Noises, they are still far from reality. To address this, we build the data collection system (see Fig.~\ref{fig:data collection}). In the system, we use a well-calibrated realistic event camera (DVXplorer) and a high fresh rate screen with a 300 fresh rate to convert RGB videos with 120 FPS to real event streams. 

First of all, we align the camera with the high refresh rate screen, ensuring that the camera aligns with the top left corner of the RGB video by using the Checkerboard alignment. Then, we replay the original videos with high temporal resolution on the high-frame screen. At the same time, we adopt the DVXplore to capture the screen to produce event data. The entire system is located in a no-light workspace, ensuring that the screen is the only light source, which can largely reduce or even eliminate the impact of screen reflection. Events captured by this pipeline could contain realistic features such as irregular noises, we illustrate this point in Fig.~\ref{fig:ev_vis}. In this figure, the left side is the input video, and the right upper is the synthetic event data, which are clean and regular. The right lower panel is the event stream generated by our system, which is dense, irregular, and noisy. There are obvious differences between synthetic and real events. Furthermore, we record the original videos again with a low temporal rate (1/5 of the original temporal resolution) because the Active Perception Sensors (APS, which is armed on some premium-version event cameras and used to capture colored images) of the event camera usually have a low frame rate, and we want to simulate the phenomenon.  
The proposed data collection system is cheap and efficient and can be used to convert any RGB video into the corresponding realistic events. The advantages of the setup include but are not limited to 1. One does not have to go outside to find various scenes to create a high-quality dataset. In contrast, they can fully utilize rich RGB video resources on the Internet 2. Overcome the low fidelity of the event camera simulator 3. One does not require a premium-version event camera (with an APS sensor), instead, our pipeline only needs a fundamental version event camera, such as DVXplorer, which is cheaper. 

\subsection{Detailed Network Architectures}
\label{Sec.architecture}

In this section, we introduce the detailed architectures and parameter selections. The networks include the deformation network to transform canonical 3D Gaussians, the PoseNet to map time to generate continuous poses, and the optical flow estimator to generate the event-based optical flow. The deformation network is learned using an MLP network $\mathcal{F}_\Phi$. This deformation network transforms the canonical position, rotation, and scale to the corresponding value given the target timestamp. The MLP receives the input and passes it to an 8-layer fully connected layers that employ the ReLU function as activation and feature 256-dimensional hidden layers and outputs a 256-dimensional feature vector. The vector is then passed through three additional prediction heads to predict the position, rotation, and scale of the 3D Gaussians. Similar to NeRF, there is a skip connection between the input feature vector and the fourth layer. Unlike DENeRF \cite{ma2023deformable}, which uses an 8-layer MLP, a more lightweight MLP is utilized in our proposed method. Our PoseNet architecture only contains the sinusoidal encoder and a 2-layer MLP to map time $t$ to translation and rotation speed ($v,w$). The network receives normalized time $t \in \mathbf{R}$ as input and output ($v,w$) following Rodrigues's formula.

The architecture of the event-based optical flow estimator is very similar to the U-Net networks. The framework receives the event spatial-temporal voxel grid as input and consists of the stridden convolution encoder, two residual block layers, and the upsample convolution decoder with skip connections to the corresponding encoder layer. We visualize the network structure in Fig.\ref{fig:evflownet}. The monocular event stream passes the downsample convolution encoders. The tanh function is applied as the activation function, and the features are passed to the residual blocks and then upsampled four times using the nearest neighbor resampling for the flow estimation. 

\subsection{The Efficiency of LoCM}
\label{sec:efficiency}
This section introduces our finetuning setting details and efficiency. We implement the LoCM finetuning by performing unsupervised adaptation of the event flow estimator for exactly 3 epochs on each scenario prior to Deformable 3DGS training, requiring only approximately 5 minutes extra computation time. This process utilizes contrast maximization (CM) as the self-supervised objective and integrates LoRA modules with a rank dimension of r=16 into the pretrained flow network while keeping the original backbone weights frozen, achieving efficient optimization and enable the optical flow estimator to adapt to unseen scenes without groundtruth optical flow as supervision.

\section{Limitations}
Although a large improvement has been achieved, this work has some limitations. This approach relies on the precomputing to extract the event flow as the motion prior which is used to guide the training of the deformation field. The pertaining of the event motion estimator block will take some time and have some biases. In the future, we plan to incorporate the motion extraction block into the whole 3DGS training pipeline. In this case, we can simultaneously train the two blocks and make them benefit from each other mutually.

\section{Conclusion}
In this work, we introduce DEGS, a novel framework that effectively overcomes the challenge of reconstructing dynamic 3DGS from low-framerate RGB videos by integrating high-temporal-resolution event streams. Experimental results demonstrate that our DEGS framework achieves superior performance in dynamic 3D reconstruction by effectively leveraging complementary multimodal data through three key innovations.

We yield some significant findings during the development and evaluation of our proposed methods. First, we experimentally demonstrate that the motion information contained in event streams significantly enhances the RGB-based 4D reconstruction method. especially in fast motions with motion blur, where events play a notably complementary role to the RGB modality. Second, we conclude that although events are a form of 2D information, their inherent connection to 3D structure allows our method to more effectively supervise the fine-grained 3D geometries. Third, we validate that fine-tuning the model via LoRA not only enables adaptation to out-of-distribution scenes but also helps the model retain its original prior knowledge, which allows our optical flow estimator to deliver high-quality motion information across diverse scenarios. Finally, through our pipeline, we demonstrate that our proposed event data generation pipeline can convert RGB data into event information with near-lossless fidelity, and this approach is applicable to a wide variety of scenarios for data generation and the scalability of event-based methods.

\section{Acknowledgment}
This work is supported by the National Natural Science Foundation of China under Grant (62405255), the Guangdong Science and Technology Department (No.2025B1212150003), the Guangzhou Municipal Science and Technology Project (No.2023A03J0013), the GuangDong Basic and Applied Basic Research Foundation (No.2023A1515110679), and the 1+1+1: HKUST-HKUST(GZ)"1+1+1" Joint Funding Program (G042).

\bibliographystyle{abbrv-doi-hyperref}

\bibliography{main}
\end{document}